\documentclass{article}
\usepackage[final]{neurips_2019}

\usepackage[utf8]{inputenc} % allow utf-8 input
\usepackage[T1]{fontenc}    % use 8-bit T1 fonts
\usepackage{hyperref}       % hyperlinks
\usepackage{url}            % simple URL typesetting
\usepackage{booktabs}       % professional-quality tables
\usepackage{amsfonts}       % blackboard math symbols
\usepackage{nicefrac}       % compact symbols for 1/2, etc.
\usepackage{microtype}      % microtypography
\usepackage{graphicx}

\title{Embedding-Based Approaches to Hyperpartisan News Detection}
\begin{document}
\maketitle
\begin{abstract}
  In this report, I describe the systems in which the objective is to determine whether a given news article could be considered as hyperpartisan. 
  Hyperpartisan news takes an extremely polarized political standpoint with an intention of creating 
  political divide among the public. 
  Several approaches, including n-grams, sentiment analysis, as well as sentence and document representations
  using pre-tained ELMo models were used. 
  The best system is using LLMs for embedding generation achieving an accuracy of around $92\%$ over the previously best system using pre-trained ELMo with Bidirectional LSTM which achieved an accuracy of around 83\% through 10-fold cross-validation. 
  
\end{abstract}

\section{Introduction}

\subsection{Hyperpartisan News}
The rise of social media and online communication has enabled people to share information with a large audience. The mask of anonymity and the lack of regulations and quality control allow malicious users to spread fake news in a destructive speed. Hyperpartisan news is a type of fake news that is typically highly polarized (hyper-partisan), emotional, and untruthful to mislead the public. Hyper-partisan articles mimic the form of regular news articles; however, they are one-sided and highly polarized in the sense that opposing voices are either deliberately ignored or attacked (Kiesel et al., 2019). Being able to detect whether a source is hyper-partisan can bring us one step closer to solving the automated fake news detection problem.

\subsection{Related Work}
The analysis of political orientation, bias, and misinformation has attracted significant attention, especially after the 2016 US presidential election. Various machine learning approaches have been made. Pla and Hurtado (2014) proposed an SVM model for political tendency identification using features such as n-grams and part-of-speech tags. Preotiuc-Pietro et al. (2017) used a linear regression model to characterize the political groups of users through language use on Twitter. In order to solve fake news problem, knowledge-based (Lee et al., 2018) and style-based approaches were proposed (Potthast et al., 2018). Furthermore, Barron-Cedeno et al. (2019) present a publicly available real-time propaganda detection system for online news using features such as n-grams and lexicon features. Some fake news datasets are publicly available, but they are often too small to be suitable for deep learning methods.

Last year, the organizers of SemEval2019 (Kiesel et al., 2019) released a large-scale dataset with over 1M articles in Task 4, Hyper-partisan News Detection, where data are labeled with the tagset {hyperpartisan, not-hyperpartisan}. Considering that many previous works have not utilized deep learning methods for hyperpartisan detection, one of our goals is to narrow down this gap by exploring how well deep learning can handle this task. Feed-forward neural networks and  Convolutional Neural Networks (CNN) were using for various types of features including sentence embeddings, n-grams, and sentiment and emotion features. 

The rest of the report is organized as follows: data and task definition are described in Section 2, Section 3 describes
our methods, followed by experiments and results in Section 4. Finally, I discuss future work and conclude this report in Section 5.

\section{Data}
\label{gen_inst}

In this report, the data used is provided by SemEval2019 task 4 (Kiesel et al., 2019). The task is set up as a binary classification problem where news articles are labeled with the target \{hyperpartisan, not-hyperpartisan\}. 

Two types of dataset are provided, pertaining how labels were obtained. The first dataset (\emph{by-article} corpus) has 1,273 articles, each labeled manually by 3 annotators at the article level (Vincent and Mestre, 2018). Out of the 1,273 labeled articles, only 645 were released by the organizer (238 hyperparsitan and 407 non-hyperpartisan), whereas the other 628 (50\% hyperpartisan and 50\% not) were reserved private for the evaluation during the competition. One of the major challenges is the range of article sizes. The maximum, mean, and minimum numbers of tokens in the by-article set are: 6470, 666, 19 respectively, making it difficult to directly input word representations as features to the neural network. As a compromise, sentence representations are used for each article by averaging the word embeddings of the corresponding sentence. 

The second, larger dataset (\emph{by-publisher} corpus) contains 754,000 articles which were automatically labeled based on a categorization of the political bias of news publishers. This dataset was split into a training set of 600,000 articles and a validation set of 150,000 articles. At first, I expected the use of by-publisher training data can help classify the by-article set, but it turned out that though much larger than the by-article set, its labels contain significant amount of noisy introduced by the labeling program (Kiesel et al., 2019). The utilization of the by-publisher set seems more difficult than expected and therefore I decided not to use it.

\section{Methodology}
\label{headings}

\subsection{Features}
\subsubsection{N-Grams}
The first obvious approach for the task was employing classic ML classifiers using the simple n-gram features. A baseline system is built using unigram and bigram features over a range of term frequency cutoff. The performance peaked when the cutoff is near 10 - 15. A variety of classifiers were tested including Logistic Regression, SVM, Random Forest, and Gradient Boosting Tree. 

\subsubsection{Emotionality (Sentiment Analysis)}
Valence (sentiment) and polarity analysis is another approach that seems suitable for the task assuming that hyperpartisan news typically involves a high intensity of valence, subjectivity, and polarity because of its biased nature. The following features were used: polarity, subjectivity; and positive, negative and neutral scores, which are obtained from textblob and NLTK text analysis. 

\subsubsection{Word Embeddings}
Word embeddings are commonly used in many NLP tasks in recent years because they are found to be useful representations of words and often lead to better performance. The input to neural networks is a set of pretrained text representations such as Word2Vec (Mikolov et al., 2013), Glove (Pennington et al., 2014), ELMo (Peters et al., 2018), or BERT (Devlin et al., 2018). In our system ELMo embeddings were used which have the advantage of modeling polysemy and morphological features that word-level embeddings could omit. Further, articles are represented as a sequence of sentence embeddings by averaging the word embeddings of those sentences. Considering the range of article size, only the first 250 sentences of an article and the first 250 tokens of a sentence was used for each article representation to account for the 512 sequence limit with pretrained BERT models. The Accuracy is 0.81. However, this may not be a good embedding model due to restriction of the sequence length.

\subsection{Classifiers and Model Architecture}

\subsubsection{Convolutional Neural Networks}
Emoloying convolving filters over neighboring words to encode information about the article seems to be another good idea. Our CNN system consists of 5 parallel convolutional layers with filter sizes 2, 3, 4, 5, 6. After each convolutional layer follows a ReLU activation function, batch normalization layer, and max pooling. The 5 parallel outputs are concatenated and fed into a fully connected layer with a sigmoid non-linearity. 

\subsubsection{Long Short Term Memory Networks (LSTMs) }
Long Short Term Memory Networks (Hochreiter and Schmidhuber,
1997) have been successfully applied to many text classification problems. RNNs are good summarizer of sequantial information such as language yet suffer from gradient issues. LSTMs solve this issue to some extent. A variation of LSTM is Bidirectional LSTMs (Bi-LSTMs), which summarizes information from both left to right and vice versa. The accuracy scores for LSTMs and Bi-LSTMs are available in table 3.

\subsubsection{Other Variety of Classifiers}
A variety of traditional machine learning classifiers using n-grams or emotionality scores as input, including Logistic Regression, SVM, Random Forest, Gradient Boosting Tree, and their ensembles were used. 

\section{Experiments and Results}

\subsection{N-gram Models}
Table 1 shows the results of our system using unigrams and bigrams with a frequency cutoff 12. Numbers appear in articles are all replaced by a token <num>. Gradient Boosting Trees turns out to be the winner, achieving an 0.775 average accuracy over 10-fold cross validation. In the second place is the ensemble of the four classifiers, indicating that the prediction of all the four models are almost identical and there is no benefit from ensembling. 

\begin{table}
  \caption{N-grams Models}
  \label{sample-table}
  \centering
  \begin{tabular}{lll}
    \toprule
    \cmidrule(r){1-2}
    Models              & Accuracy      \\
    \midrule
    Logistic Regression & 0.7535     \\
    SVM                 & 0.7380      \\
    Random Forest       & 0.7534    \\
    XGBoost             & 0.7752    \\
    Ensemble            & 0.7690    \\
    \bottomrule
  \end{tabular}
\end{table}

\subsection{Sentiment Analysis}
Table 2 shows the results of our system using polarity, subjectivity, and valence scores. The results are much worse than n-gram systems. There could be three reasons for the poor performance. First, the methods used to generate the scores were not very optimal. Second, to fully leverage this methods, more hand-crafted lexicon-based features were used.

\begin{table}
  \caption{Sentiment Analysis Models}
  \label{sample-table}
  \centering
  \begin{tabular}{lll}
    \toprule
    \cmidrule(r){1-2}
    Models              & Accuracy      \\
    \midrule
    Logistic Regression & 0.6372     \\
    SVM                 & 0.6310      \\
    Random Forest       & 0.6881    \\
    XGBoost             & 0.7021    \\
    Ensemble             & 0.6972    \\
    \bottomrule
  \end{tabular}
\end{table}

\subsection{Neural Networks}
Table 3 illustrates the average accuracy over 10-fold cross validation obtained on the systems of CNN and LSTMs. Dropout is used for feedforward networks and LSTM and batch normalization is used for feedforward networks and CNN. The four systems are comparable but LLM significantly outperforms the others. 

\begin{table}
  \caption{Neural Networks}
  \label{sample-table}
  \centering
  \begin{tabular}{lll}
    \toprule
    \cmidrule(r){1-2}
    Models              & Accuracy      \\
    \midrule
    CNN             & 0.8095      \\
    LSTM            & 0.8218    \\
    Bi-LSTM         & 0.8372    \\
    LLM (Gemma3)    & 0.9144    \\
    LLM (llama)     & 0.9262    \\
    \bottomrule
  \end{tabular}
\end{table}

\subsection{Overfitting Investigation}
Overfitting has been a problem throughout the experiments. This is reasonable considering that the size of training size is only 645. Thus,  experiments were conducted on how the number of training examples will influence performance. Figure 1 illustrates the result, suggesting that the system could perform even better if given a larger training set. 

\begin{figure}[htp]
    \centering
    \includegraphics[width=4cm]
    {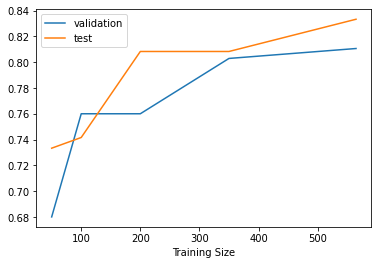}
    \caption{Training size vs. Accuracy}
    \label{fig:galaxy}
\end{figure}

\subsection{LLMs}
Recently, LLMs are being used to generate embeddings. These embeddings capture more nuanced representation in identifying hyperpartisan news. Unlike traditional word embeddings, such as Word2Vec or GloVe, which rely primarily on local context and static representations, embeddings generated by large language models incorporate dynamic contextual information. This allows for better detection of subtle linguistic cues, ideological framing, and rhetorical strategies commonly present in hyperpartisan content. As a result, these embeddings improve the performance of downstream classification models, enhancing their ability to distinguish between balanced reporting and biased narratives. 

Additionally, leveraging embeddings from LLMs enables transfer learning across datasets, making models more robust in real-world applications where labeled partisan data may be scarce. 

The embeddings generated using Llama‑4‑Scout‑17B‑16E‑Instruct improved the classifier accuracy to 92.62\% while the embeddings generated using gemma3 models improved the classifier accuracy to 91.44\%

\section{Discussion and Conclusion}
In this report, I present the systems on hyperpartisan news detection. I trained a set of models using a variety of features including n-grams, sentiment scores, and word embeddings. The Bi-LSTM system using ELMo embedding acquired a 83.72\% accuracy through 10-fold cross validation without much hyperparameter tuning. The overfitting problem investigation concludes that potential of deep learning methods has not been fully exploited due to the small training set. Lastly, the embeddings generated using LLMs achieved an accuracy of 92.62\%. Embeddings generated using LLMs improved the classifier accuracy over the previously built system using ELMo models.

\section*{References}

\small

[1] Johannes Kiesel, Maria Mestre, Rishabh Shukla, Emmanuel Vincent, Payam Adineh, David Corney, Benno Stein, and Martin Potthast. 2019. SemEval–2019 Task 4: Hyperpartisan News Detection. In {\it Proceedings of The 13th International Workshop on Semantic Evaluation}. Association for Computational Linguistics.

[2] Ferran Pla and Lluis-F Hurtado. 2014. Political tendency identification in twitter using sentiment analysis techniques. In {\it Proceedings of COLING 2014, the 25th international conference on computational linguistics}: Technical Papers, pages 183–192.

[3] Daniel Preotiuc-Pietro, Ye Liu, Daniel Hopkins, and Lyle Ungar. 2017. Beyond binary labels: political ideology prediction of twitter users. In {\it Proceedings of the 55th Annual Meeting of the Association for Computational Linguistics} (Volume 1: Long Papers), volume 1, pages 729–740.

[4] Nayeon Lee, Chien-Sheng Wu, and Pascale Fung. 2018. Improving large-scale fact-checking using decomposable attention models and lexical tagging. In {\it Proceedings of the 2018 Conference on Empirical Methods in Natural Language Processing}, pages 1133–1138.

[5] Martin Potthast, Johannes Kiesel, Kevin Reinartz, Janek Bevendorff, and Benno Stein. 2018. A Stylometric Inquiry into Hyperpartisan and Fake News. In {\it 56th Annual Meeting of the Association for Computational Linguistics} (ACL 2018), pages 231–240. Association for Computational Linguistics

[6] Alberto Barron-Cedeno, Giovanni Da San Martino, Israa Jaradat, and Preslav Nakov. 2019. Proppy: A system to unmask propaganda in online news. In {\it Proceedings of the Thirty-Third AAAI Conference on Artificial Intelligence} (AAAI’19), AAAI’19, Honolulu, HI, USA.

[7] Emmanuel Vincent and Maria Mestre. 2018. Crowdsourced Measure of News Articles Bias: Assessing Contributors’ Reliability. In {\it Proceedings of the 1st Workshop on Subjectivity, Ambiguity and Disagreement (SAD) in Crowdsourcing}, pages 1–10.

[8] Tomas Mikolov, Ilya Sutskever, Kai Chen, Greg S Corrado, and Jeff Dean. 2013. Distributed representations of words and phrases and their compositionality. In {\it Advances in neural information processing systems}, pages 3111–3119.

[9] Jeffrey Pennington, Richard Socher, and Christopher Manning. 2014. Glove: Global vectors for word representation. In {\it Proceedings of the 2014 conference on empirical methods in natural language processing (EMNLP)}, pages 1532–1543.

[10] Matthew E Peters, Mark Neumann, Mohit Iyyer, Matt Gardner, Christopher Clark, Kenton Lee, and Luke Zettlemoyer. 2018. Deep contextualized word representations. {\it arXiv preprint arXiv:1802.05365.}

[11] Jacob Devlin, Ming-Wei Chang, Kenton Lee, and Kristina Toutanova. 2018. BERT: Pre-training of Deep Bidirectional Transformers for Language Understanding. {\it CoRR}, abs/1810.04805.

[12] Sergey Ioffe and Christian Szegedy. 2015. Batch normalization: Accelerating deep network training by reducing internal covariate shift. {\it arXiv preprint arXiv:1502.03167.}

[13] Nitish Srivastava, Geoffrey Hinton, Alex Krizhevsky, Ilya Sutskever, and Ruslan Salakhutdinov. 2014. Dropout: a simple way to prevent neural networks from overfitting. {\it The Journal of Machine Learning Research}, 15(1):1929–1958.

[14] Sepp Hochreiter and J\"urgen Schmidhuber. 1997. Long short-term memory. {\it Neural computation}, 9(8):1735–1780.

[15] Hugo Touvron, Thibaut Lavril, Gautier Izacard, Xavier Martinet, Marie-Anne Lachaux, Timothée Lacroix, Baptiste Rozière, Naman Goyal, Eric Hambro, Faisal Azhar, Aurelien Rodriguez, Armand Joulin, Edouard Grave, and Guillaume Lample. 2023. LLaMA: Open and Efficient Foundation Language Models. \textit{arXiv preprint arXiv:2302.13971}.

[16] Gemma Team, Aishwarya Kamath, Johan Ferret, Shreya Pathak, Nino Vieillard, Ramona Merhej, Sarah Perrin, Tatiana Matejovicova, Alexandre Ramé, Morgane Rivière, \textit{et al.} 2025. Gemma 3 Technical Report. \textit{arXiv preprint arXiv:2503.19786}.

\end{document}